\documentclass[runningheads,a4paper]{llncs}

\usepackage{amssymb}
\usepackage{graphicx}

\usepackage{url}
\urldef{\mailsa}\path|{pan ,yakovlev}@isa.ru|    
\newcommand{\keywords}[1]{\par\addvspace\baselineskip
\noindent\keywordname\enspace\ignorespaces#1}

\begin{document}

\mainmatter  

\title{Behavior and path planning for the coalition of cognitive robots in smart relocation tasks}

\titlerunning{Behavior and path planning for the coalition of cognitive robots}

\author{Aleksandr I. Panov \and Konstantin Yakovlev}
\authorrunning{Aleksandr I. Panov \and Konstantin Yakovlev}

\institute{Federal Research Center ``Computer Science and Control'' of RAS,\\
pr. 60-letiya Octyabrya 9, 117312 Moscow, Russia\\
}

\toctitle{Behavior and path planning for the coalition of cognitive robots in smart relocation tasks}
\tocauthor{Aleksandr I. Panov, Konstantin Yakovlev}
\maketitle

\begin{abstract}
In this paper we outline the approach of solving special type of navigation tasks for robotic systems, when a coalition of robots (agents) acts in the 2D environment, which can be modified by the actions, and share the same goal location. The latter is originally unreachable for some members of the coalition, but the common task still can be accomplished as the agents can assist each other (e.g. by modifying the environment). We call such tasks smart relocation tasks (as the can not be solved by pure path planning methods) and study spatial and behavior interaction of robots while solving them. We use cognitive approach and introduce semiotic knowledge representation --- sign world model which underlines behavioral planning methodology. Planning is viewed as a recursive search process in the hierarchical state-space induced by sings with path planning signs reside on the lowest level. Reaching this level triggers path planning which is accomplished by state of the art grid-based planners focused on producing smooth paths (e.g. LIAN) and thus indirectly guarantying feasibility of that paths against agent's dynamic constraints.
\end{abstract}

\section{Introduction}

In pursuit of higher autonomy degree of modern robotics systems researchers combine various methods and algorithms of Artificial Intelligence, Cognitive Science, Control Theory into so-called Intelligent Control Systems (ICS) \cite{Albus2002,Yoo2015}. These systems are the collections of software modules automating robot behavior and are conventionally organized in a hierarchical fashion. Usually three levels of control --- strategic, tactical and reactive (or named in another way but still bearing the same sense) --- are distinguished \cite{Emelyanov2015}. In this work, we address the planning problem and examine planning methods on both strategic and tactical levels and their interaction. On strategic level, we assume that there exist a description of situations and goals, and the search space induced by such descriptions is processed to produce a valid plan. Typically in AI planning \cite{Ghallab2004} first-order logic is used to model the world as well as specialized first-order logic languages --- PDDL and it's derivatives --- are used to formalize robot's actions \cite{Ghallab1998,Fox2003}. Planning relying on these models and languages is known as task planning. In our work a new formalism --- sign world model --- is introduced which is based on cognitive theories, takes into account results of recent cognitive and neurophysiologic research and thus make robot's behavior more human-like, robust and versatile. We refer to planning based on sign world model as to behavior planning. As for the tactical level of control system, it deals mainly with navigation tasks so planning is considered in spatial (geometrical) sense and is aimed at finding a path (feasible trajectory) for the robot. Both planning activities --- behavior planning and path planning --- despite the common term involved in their names utilize different models and algorithms and commonly are studied independently. In the present work, we study them as a part of coherent framework. One should say, that there exists a limited number of the approaches of task (not --- behavior) and path planning integration, see \cite{Karlsson2012,Abdo2012} for example. These approaches mainly examine some aspects of task and path planning integration when there is a single robot interacting with the environment. In our work, we study the behavior of the coalition of robots and how integration of planning activities on both strategic and tactical layer can affect such behavior. We examine navigation tasks in 2D world which can be transformed by the robots' actions. More precisely, we investigate the case when unsolvable for some member of the coalition problem can be solved if other robots alter their plans and assist each other. We call such tasks --- smart relocation tasks (as they can not be accomplished by path planning only methods).

The latter of the paper is organized as follows --- in Section~\ref{works} related works regarding task and path planning are discussed. Section~\ref{case} contains the description of the smart relocation task we are interested in. Novel world modeling formalism, which utilizes cognitive approach, is introduced in Section~\ref{knowledge} and in Section~\ref{behavior} planning method based on this formalism is described. Suggested path planning approach is discussed in Section~\ref{path}. Model example in studied in Section~\ref{example}.

\section{Related works}\label{works}
\subsection{World modeling and behavior planning}

Behavior (task) planning is the main objective of control systems based on cognitive architectures. Well-known SOAR \cite{Laird2008,Laird2012} system is considered as the industry standard in this area. In SOAR, as well is in majority of other cognitive control systems, agent's memory is separated into the long term memory, the short term memory and the memory of estimates. Objects, situations and goals are represented in the short term memory in the form of attribute descriptions. The long term memory contains transitions (operators) between short term memory states and is represented by AI rules \cite{Nilsson1998}.
 
Agent's planning procedure in SOAR consists of a sequence of decisions, where the aim of each decision is to select and apply an operator in service of the agent's goals. The simple decision circle contains five steps: encode perceptual input, activate rules to elaborate agent's state (propose and evaluate operators) in parallel, select an operator, activate rules in parallel that apply the operator and then process output directives and retrievals from long-term memory.

Similar knowledge representation and planning method are implemented in other cognitive control systems. In the Icarus project \cite{Langley1997,Langley2006} the division of the long term memory into conceptual and action memories was introduced. Planning procedure of Icarus relies on recursive action decomposition up to low level actions, called skills. Skill sequence is executed when start situation is satisfied in short term memory. In Clarion \cite{Sun1994,Sun2006} some rules of action choosing are based on neural networks. Thus knowledge representation in Clarion contains not only explicit (attributive) component but also an implicit one. The learning process on the set of predefined precedents is the distinctive feature of this system.

Modern algorithms of behavior (task) planning use so called STRIPS description of planning domain \cite{Fikes1971}. One of the main directions in task planning is the development of special graph structures encoding both state descriptions and state transitions for further search. The first algorithm using graph representation was Graphplan \cite{Blum1997}. Graphplan search procedure is executed on the special layered compact planning graph and returns a shortest-possible partial-order plan or state that indicates the absence of the valid plan.

Further research in this area was concentrated on development of specialized search algorithms for these graph structures. For example in the Fast Forward (FF) \cite{Hoffmann2001} and the Fast Downward (FD) planning systems \cite{Helmert2006} heuristic search is used. These planners are aimed at solving general deterministic planning problems encoded in the propositional fragment of PDDL description \cite{Fox2003} and search the state space in the forward direction. FF, FD and other widespread PDDL-based planners use the propositional representation with special implicit constraints being considered in some cases. For example, FD planner computes its causal graph  heuristic function taking these implicit constraints into account as well as using hierarchical decompositions of planning tasks.

Another remarkable heuristic planning system is LAMA \cite{Richter2010}. It uses pseudo-heuristic derived from landmarks --- propositional formulas that must be true in every solution of a planning task. The LAMA system builds plan using finite domain rather than binary state variables as in the FF planner.

One should note that the propositional language for task description is not relevant to many real problems. Thus extensions of the language and development of hybrid planning domains is appealing research area. For example, UPMorphi universal planner \cite{Della2012} is capable of reasoning with mixed discrete and continuous domains and fully respect  the semantics of PDDL+ \cite{Fox2006}. UPMorhi performs universal planning on some initial discretization and checks the correctness of the result. If the validation fails, discretization is refined and algorithm is reinvoked. 

All of the abovementioned and other existing planners are not suitable for the cooperative behavior (task) planning. Special knowledge representation such as MA-PDDL \cite{Kovacs2012} should be used in this case. These representations and planners based on them should solve symbol grounding problem \cite{Harnad1990} and support goal-setting and role distribution procedures. Such requirements can be met by the sign representation, which is based on the models of cognitive functions \cite{Osipov2014} and neurophysiological studies of the cognition process \cite{Edelman1987,Ivanitsky1997}. We will use this approach to realize communication protocol for cooperative planning and providing a link between the symbolic models and sensor (low level) data.

\subsection{Spatial modeling and path planning}

Traditionally in artificial intelligence and robotics path planning is viewed as a graph search process. Agent's knowledge about the environment is encoded into the graph model and the search for a path on that graph is performed. Typically, graph's vertices correspond to the locations an agent can occupy and edges --- to the trajectories it can traverse (for example --- straight sections or curves of predefined lengths and curvatures). Weighting function, which assigns weights to the edges, is commonly used to quantitatively express any characteristics of the corresponding trajectories (length, energy cost, risk of traversing etc.). So to plan a path one needs to a) construct a graph model out of the environment description available to an agent b) find a (shortest) path on that graph.

The most widespread graph models used as the spatial world model of an agent are Visibility Graphs \cite{Lozano1979}, Voronoi Diagrams \cite{Bhattacharya2008}, Navigation Meshes \cite{Kallmann2010}, Regular Grids \cite{Yap2002} etc. Each of them needs it's own algorithm to be executed to transform raw information about the environment to the model. In case environment is compound of the free space and the polygonal obstacles (the most widespread case), two graph models are typically used --- visibility graphs and regular grids. Constructing visibility graph is computationally burdensome and each time goal position changes additional calculations should be performed to add corresponding edges to VG \cite{Wooden2006}. Algorithm of grid construction is much more simple --- it's complexity is a constant in respect of number of obstacles' vertices and edges, and no additional calculations should be made when goal or start position alters. So, grids can be referred to as simple yet informative graph models, and in most cases it is the grids that are used for path planning. Another reason grids are so widespread is that new knowledge on the environment gained via sensor information processing can be easily integrated into them \cite{Elfes1989} without the necessity to re-invoke graph construction algorithm, which significantly saves computational resources.

After the graph is constructed, the search for a path is performed (typically, the shortest path is targeted). There exist a handful of algorithms for that: Dijkstra \cite{Dijkstra1959}, A* \cite{Hart1968} --- which is the heuristic modification of Dijkstra, and many of their derivatives: R* \cite{Likhachev2008}, Theta* \cite{Nash2007}, JPS \cite{Harabor2011}, D* Lite \cite{Koenig2000} etc. Some of these algorithms are specifically tailored to grid path planning (like JPS or Theta*) some work on any graphs (D* Lite, R*) with A* and Dijkstra being the most universal ones (and the most computationally ineffective while solving practical tasks as well).

If we are talking not about an abstract agent, which can move in any directions, with any speed and acceleration, and stop instantly, we need to take into account agent dynamic constraints while searching for a path. Common way to consider these constraints is to incorporate them somehow into the graph model or, which is nearly the same, into the search space --- see \cite{Kuwata2009} for example. The main problem here is that the search space becomes orders of magnitude times larger, especially when an agent exhibits rather complex dynamics (for example - multirotor UAV).  Another problem here is that admissible, monotone, well-informed heuristics utilized to guide the search can be easily introduced only for the spatial-only search spaces, which is not the case anymore. Summing up the above mentioned one can claim that it may be beneficial to stay within spatial-only search spaces but search for such paths that indirectly guarantee feasibility against the agent's dynamic constraints, e.g. smooth paths not containing sharp turns.  One of the recently introduced approach in this area, is planning for angle-constrained paths \cite{Kim2014}. We believe that this approach is very promising and suggest using LIAN algorithm \cite{Yakovlev2015} for agent's path planning. To the best of our knowledge it's the only angle-constrained path planning algorithm which is sound and complete (in respect to it's input parameters). 

When talking about coalitions of agents and multi-agent grid path planning the most well-studied problem is resolving spatial conflicts for groups of agents with primitive dynamics, e.g. agents that can move from an arbitrary grid cell to any of it's 8 adjacent neighbors and stop (and later on start) moving instantaneously. There exist both sound and complete but very computationally expensive methods of solving this task \cite{Standley2010} and fast but incomplete algorithms \cite{Wang2008,Silver2006}. Much less attention is paid to the problem of agents spatial interaction when planning for a path --- a problem which will be addressed in our work in more details.

\subsection{Summary}
Currently existing cognitive control systems and PDDL-based planners don't consider some important features of the planning problem in case coalition of interacting agents is involved. Dynamic formation of goals and goal sharing in the context of changing environment impose special restrictions on the knowledge representation to be used by planning systems. Necessity to divide believes of a single agent into communicable and personal parts presents another restriction. It's also worth noting that within existing task planning frameworks little attention is paid to coordination of path planning process and behavior knowledge about the environment. Regarding path planning itself one can state that grid-based path planning is the most widespread methodology as grids are simple yet informative spatial models and a handful of methods tailored to grid path finding exist. Unfortunately grid-based paths do not take into account agent's dynamic constraints while incorporating dynamic laws encodings into the  search process severely degrades overall performance (due to the enormous extension of the search space). So it can be beneficial to stay within spatial only search spaces but search for a specific, geometrically constrained class of paths and thus indirectly guaranty path's feasibility. Further on we will present a coherent task-path planning framework which addresses all the mentioned concerns and bottlenecks.

\section{Considered case}\label{case}

Further on we will use the term agent as well as robot (robotic system) following the conventions of AI literature.

\begin{figure}[h]
	\centering
	\includegraphics[height=6cm]{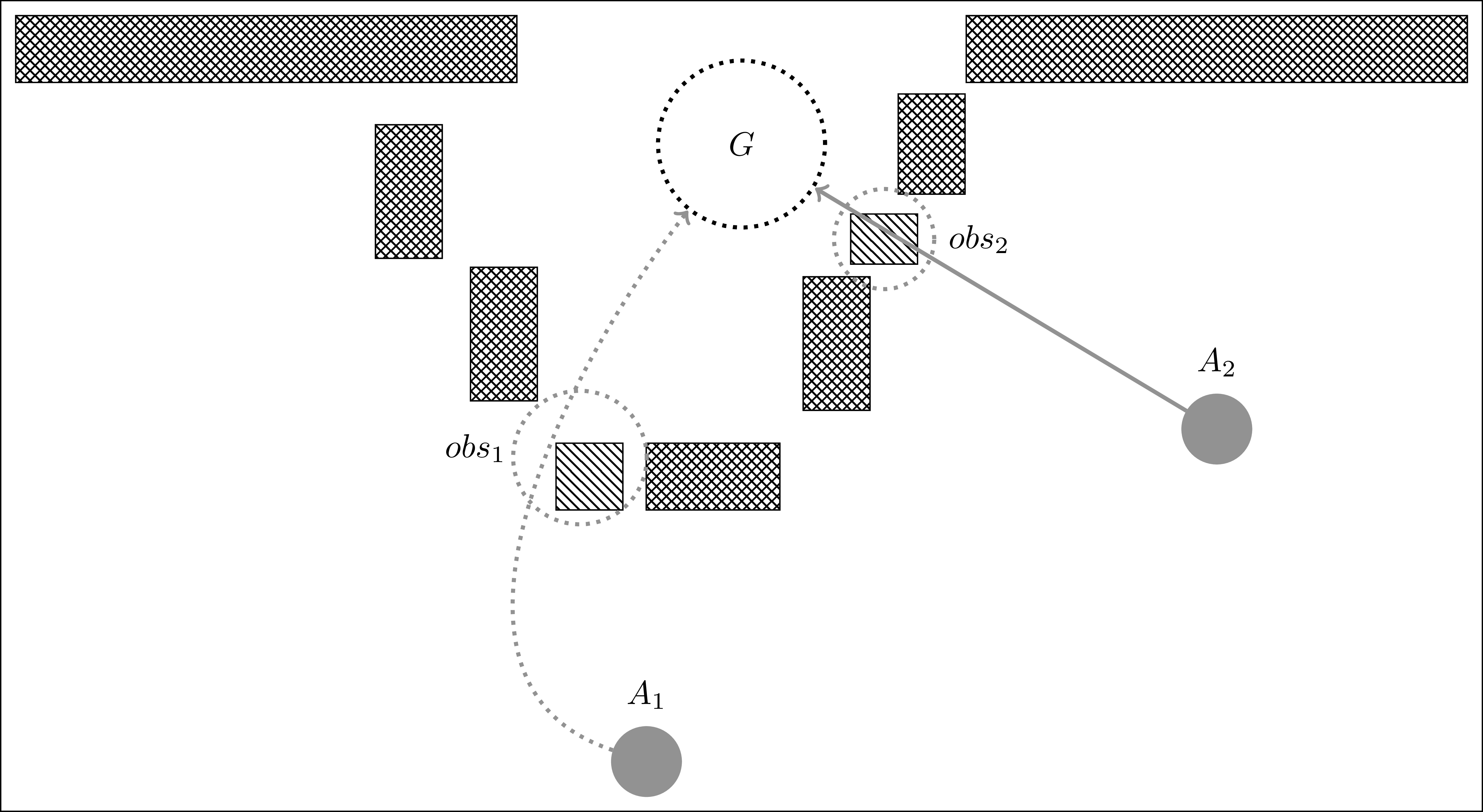}
	\caption{Considered case of the coalition relocation task: $A_1,A_2$ --- members of the coalition, $G$ --- the goal area, $obs_1, obs_2$ --- obstacles of the $ot_1$ type (inclined lines blocks) destroyable by agent $A_2$.}
	\label{fig:case}
\end{figure}

We consider the following task. The coalition of agents $A=\{A_1,\dots, A_N\}$ acts in the static  environment (workspace) which is the rectangular area of 2D Euclidean space $U: x_{min} \leq x \leq x_{max}, y_{min} \leq y \leq y_{max}$. $U$ is comprised out of the free space $U_{free}$ and the obstacles $U_{obs}=\{obs_1,\dots,obs_M\}$. Each obstacle is a polygon defined by the set of it's vertices' coordinates $obs_i=\{p_{i1}, p_{i2}, \dots, p_{ij}, \dots, p_{iK_i}\}$, $p_{ij}=(x_{ij}, y_{ij})\in U$. Obstacles are additionally characterized by types: $type(obs_i)=ot_j$, $ot_j\in OT$, $OT=\{ot_1, \dots, ot_Z\}$. All agents have similar sizes and can be represented as the circles of radius $r$ in $U$ (see Fig.~\ref{fig:case}.

We suppose that the agent's movement dynamics is encoded as a set of differential equations:
\begin{equation}
	\frac{dx}{dt} = f(x,u),
\end{equation}
where $f(x,u)$ --- vector function, $x\in R^n$ --- vector of the phase coordinates, $u\in R^r$ --- control vector, $t$ --- time. Following \cite{Yakovlev2015a} we assume that given dynamic constraints can be transformed to geometry constraints, e.~g. we assume that a feasible trajectory for an agent is the angle-constrained path in $U$ which is a sequence of line segments such that the angle of alteration between two consecutive segments doesn't exceed predefined threshold $\alpha_m$.

Agent's knowledge base contains high-level representations of locations and distances as well as the mechanisms of mapping these representations to the workspace. Set of agent's actions is organized in hierarchical structure and three types of actions exist: transition actions, transforming actions (which are limited to destroying obstacles of different types) and messaging actions. We consider the case when each agent has its own planning focus containing current believes about external objects and processes. Details of the knowledge representation will be described further in Section \ref{knowledge}.

Single agent's task is reaching the predefined goal area which is the same for all other agents. This common task description for an agent includes explicit constraint that all the agents should reach the goal area (not the only one). We investigate scenarios (as depicted on Fig.~\ref{fig:case}) when some agents can not reach the goal area separately, without the assistance from the other members of the coalition. As seen on Fig.~\ref{fig:case} agent $A_1$ can not reach the goal as it's blocked by the obstacle $obs_1$ which can not be destroyed by $A_1$, while $A_2$ can alter it's plan, reach $obs_1$ first and destroy this obstacle assisting $A_1$ in accomplishing the task). We call such tasks --- smart relocation tasks.

\section{Knowledge representation}\label{knowledge}

\begin{figure}
	\centering
	\includegraphics[height=8cm]{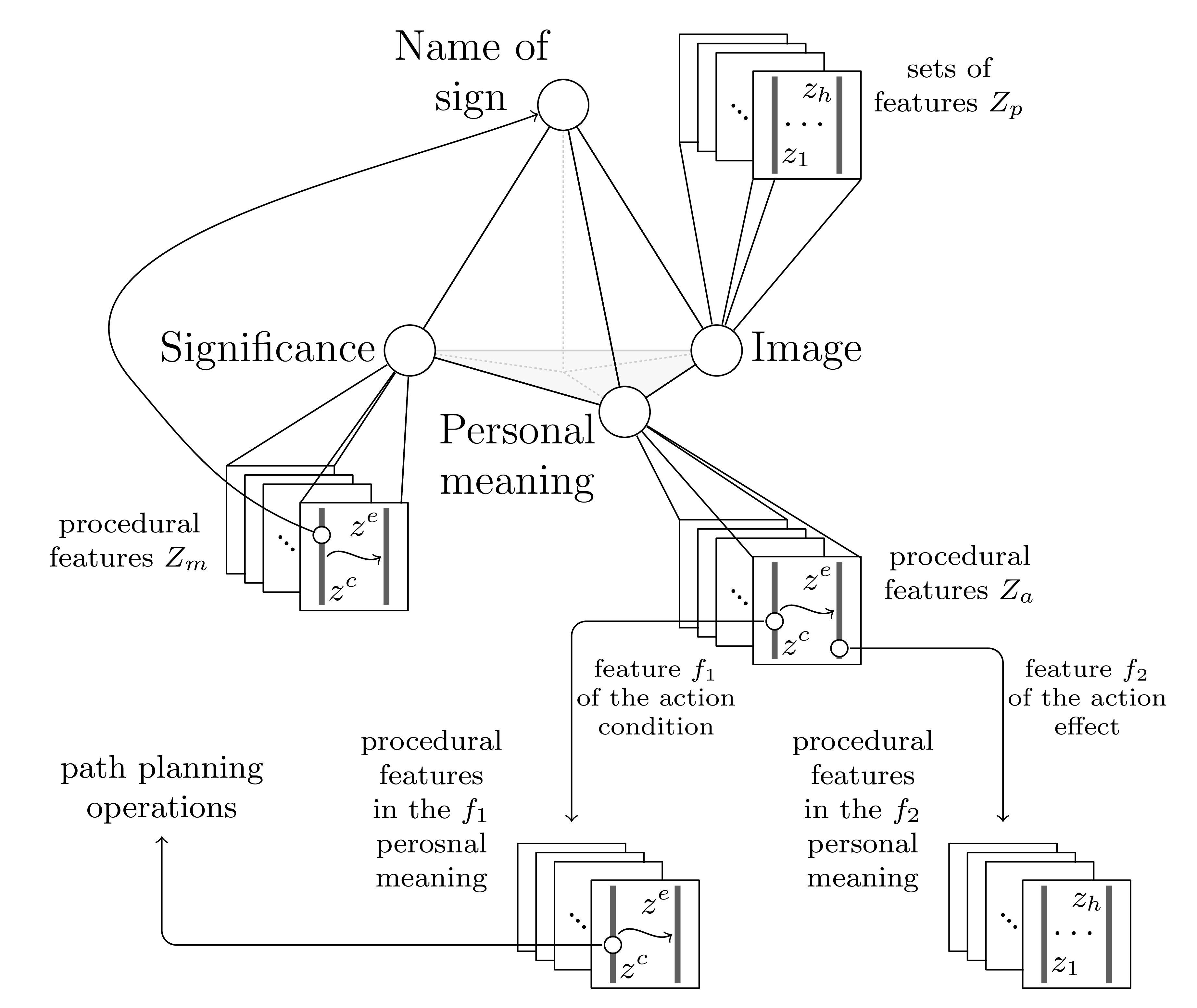}
	\caption{Structure of the sign knowledge representation.}
	\label{fig:sign}
\end{figure}

Agent's knowledge base contains descriptions of objects, processes and properties of the external environment and information about other members of the coalition. To formalize the knowledge base we use the semiotic approach where-in all above entities are mediated by sings.  Each sign is composed of a name and three components --- image, significance, personal meaning --- which are used to implement different functional steps of the planning process. Signs come with the special structured set of links to other sings and to data from inner and external sensors of the agent. We will name these links as features (see Fig.~\ref{fig:sign}).

The first component of a sign is image. Image is the set of structured sets of features specific to the mediated entity. Each structured set of features (shown as square on Fig.~\ref{fig:sign}) corresponds to particular group of characteristic properties of the entity mediated by sign and differ from the other by it's structure as well as by the features themselves. At the same time image also implements the process of recognizing the entity based on the input data. Each set of features encapsulated in image contains those features that are grouped together in the input data stream --- shown as columns inside the square on Fig.~\ref{fig:sign}. Each column contains features that together form certain part of the mediated entity description. Considering that features are links to other signs, the hierarchy of signs is formed on the set of images. Lowest level of the hierarchy consists of input data from sensors or information received from path planning operators.

The second component of a sign is significance. Significance is the set of procedural features (or causal relations) and it is used to describe characteristic actions in which mediated entity is engaged. A causal relation consists of the set of conditional features or conditions (encountered before the execution of the action) and the set of resultant features or effects (encountered after the execution of the action). Thus procedural features are the models of AI rules \cite{Nilsson1998}. Additionally, at least one feature (condition or effect) is a link to the sign possessing the significance itself. Considering that features are links to other signs, another hierarchy of signs is formed on the set of significances of mediated actions. Lowest level of the hierarchy contains elementary skills . Significance components of common signs are the same for all agents in the coalition.

Finally the third component of a sign is personal meaning and it also (like significance) describes actions involving the mediated entity. There exist a link between a causal relation of the personal meaning and a causal relation of the significance defined by the function $\Xi$. Unlike significance, personal meaning contains special type of features --- personal features --- in it's causal relations. These features mediate inner properties of the agent and replace elementary skills of the corresponding significance. Personal meaning implements the process of applying agent's actions. Lowest level of the personal meanings hierarchy is comprised by path planning operators. 

The structure of the sign corresponds to psychological models of high cognitive functions \cite{Vygotsky1986,Leontyev2009} and allows to separate generalized representation of actions that are known to all members of the coalition and specific implementation of such actions, which takes into account inner properties of the agent.

The hierarchy of signs (based on images, significances and personal meanings) serves as a tool for the input signal (low level features) recognition and for the corresponding sign actualization. An algorithm of sign recognition is simple comparison of input features with corresponding set of features predicted by upper level signal on each level of the hierarchy \cite{Osipov2014}. In this way the recognition process is bottom-up spreading of the activation in the hierarchy of features right down to levels where there is correspondence between features and signs. This algorithm models functioning of the human cortex sensor regions \cite{Mountcastle1998,George2009}.The set of activated (actualized) signs at the moment represents the agent's believe about the current environment state. Since the hierarchy of signs encodes the set of agent's actions via the procedural features all transitions between states during behavior planning are executed as top-down or bottom-up activation processes in the hierarchy. Low level procedural features include path planning operators and reaching this level while the activation (planning) process triggers path planning procedures.

\section{Behavior planning algorithm}\label{behavior}

On sign level behavior planning is realized in the situation space by $PMA$ algorithm proposed in \cite{Osipov2015}. The situation is defined as the set of signs structured in the same way as the image components (see above), e.\,g. signs are split into groups which describe different parts of the situation. Use of sign representation allows to combine believes about relationships and believes about objects and consider all properties, processes and objects in a situation as signs. Transitions between situations are implemented by casual relations contained in procedural features of significances or personal meanings in dependence the planning step. The initial situation is defined as the current observed situation, i.\,e. the current set of actualized signs. The goal situation is agent's believe about the result of the solving current problem, i.\,e. it is the set of goal signs.

In relocation tasks low level features are implemented by path planning algorithms and are considered to be personal features included in agent's personal meanings. Thus the hierarchy of procedural features in fact is the action hierarchy and low level actions are performed by the subsystem of path planning which lies beyond sign representation.

The algorithm of behavior planning is an iterative process ($PMA$-procedure) consisting of the following steps:
\begin{itemize}
	\item search of relevant significances (the $M$-step), 
	\item choose a personal meaning from the set of personal meanings corresponding to the found significances (the $A$-step), 
	\item send a message to other members of the coalition (part of the $S$-step),
	\item perform the action corresponding to the chosen personal meaning (part of the $S$-step),
	\item construct the new current situation using the set of features from the condition of performed action (the $P$-step).
\end{itemize}

Input of the $PMA$-procedure is the pair of two situations: start and final situations. On the first iteration of the algorithm the start situation is the current situation (observed by the agent) and the final situation is the goal situation. Then the $M$-step is done, e.\,g. the search on the significances of signs forming the final situation is performed. Effect parts of each significances are considered and each effect is compared to the set of signs of final situation (such comparison is valid as each effect is a set of features linked to corresponding signs). After the comparison is done such significances are selected which effect features matches maximum number of final signs.

The $A$-step is execution of $\Xi$ procedure which associates the set of procedural features of the selected significances  with the procedural features of personal meanings. Then the selection of one procedural feature of personal meaning is done. Transition to the $P$-step occurs. The $P$-step of the algorithm is combining of all features included in conditions of personal meaning features. After $P$-step the new iteration of $PMA$-procedure is executed with the same start situation and the new formed final situation.

If new situation is the subset of start situation (the first argument of $PMA$-procedure) then in some cases $S$-step occurs depending on cognitive qualities (parameters of planning process) of the agent. $S$-step results in defining the new goal situation as a result of the action (rule) application. This application adds to current situation description features from the image component corresponding to procedural features of personal meanings defined on the $A$-step. The activation process spreads top-down in the procedural features hierarchy (see Section~\ref{knowledge}). In some cases activation process reaches the level where path planning operators are included in personal features, so path planning is started. If path planner returns success $PMA$-procedure also ends with \textit{success} --- behavior plan including valid path of relocation to the goal area is constructed. If path planner returns \textit{failure} along with the coordinates of blocking obstacle (see below) then the new features (corresponding to the identified obstacle) are added to the description of the current situation and the $PMA$-procedure is repeated.

\begin{figure}
	\centering
	\includegraphics[height=6.5cm]{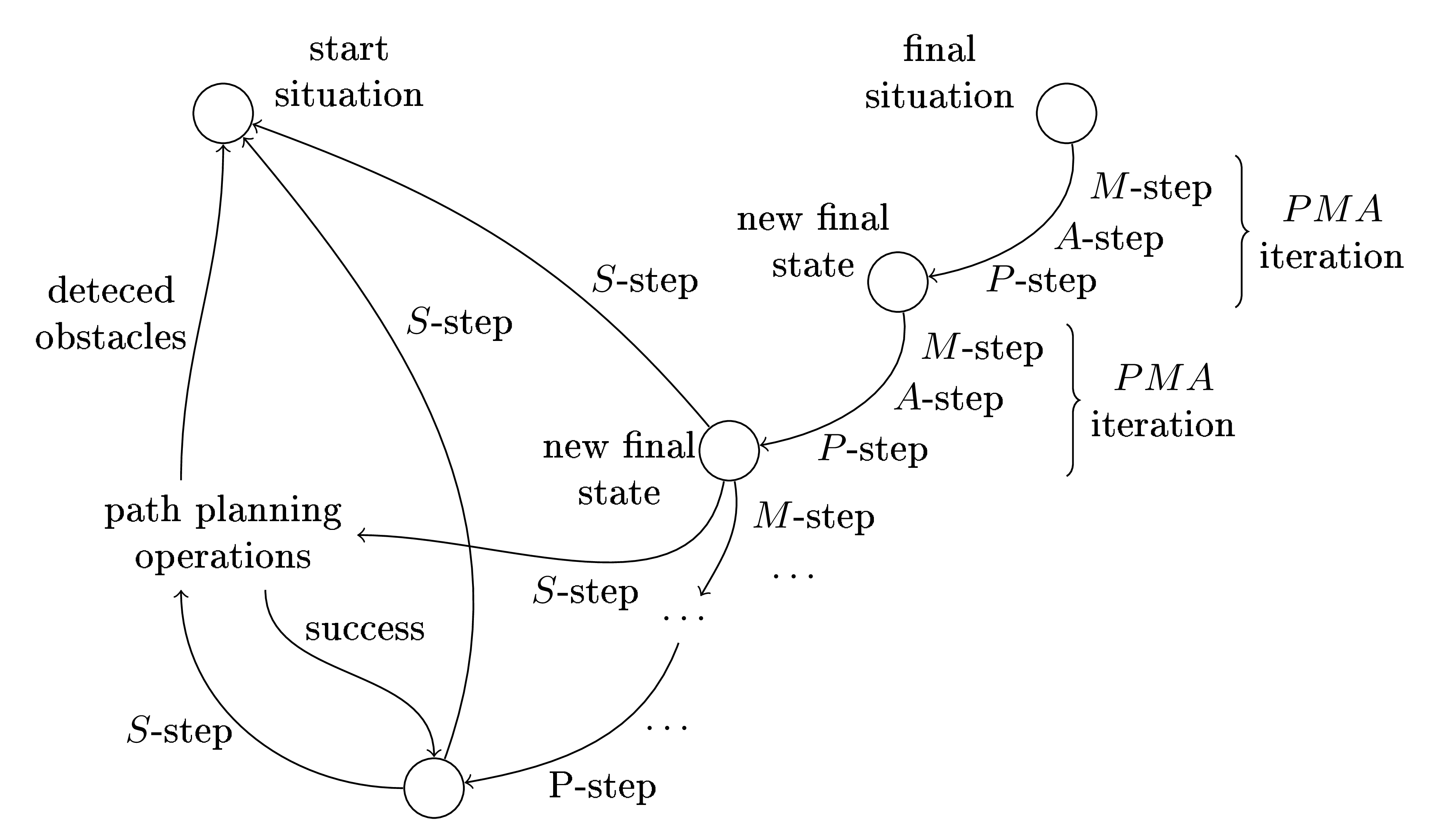}
	\caption{Schema of behavior planning $PMA$-procedure}
	\label{fig:bplan}
\end{figure}

All changes arising in the current observed situation (for example, emerging unaccounted obstacle detected by path planning process or the new task received from other member of the coalition) triggers the re-execution of $PMA$-procedure with new start or final situations.

With new sign representation of agent's knowledge about environment and its own qualities we can describe and implement meta-cognitive regulation functions of the agent behavior. These functions are realized by rule (mental action) application mediated by personal meanings and significances during the selection of $PMA$-procedure parameters. This regulation process executes some rules that change personal features of the agent implemented parameters of planning and recognition processes.

All members of the coalition have signs that mediate both objects of external environment and other members of the coalition. The significances of these signs include agent's knowledge about actions available for corresponding agents. The personal meanings of these signs include actions by sending them a communication messages.
The constructed plan of behavior can contain personal meanings of signs corresponding to other members of the coalition. In this case a message with description of the significance obtained by the inverse procedure $\Xi$ is sent to the corresponding member. This message plays the role of the new task for this agent and triggers its $PMA$-procedure re-execution. Thus the common plan of the task resolving includes all sub-plans and all goal-setting messages of all members of the coalition.

\section{Path planning algorithm}\label{path}

\begin{figure}
	\centering
	\includegraphics[height=6cm]{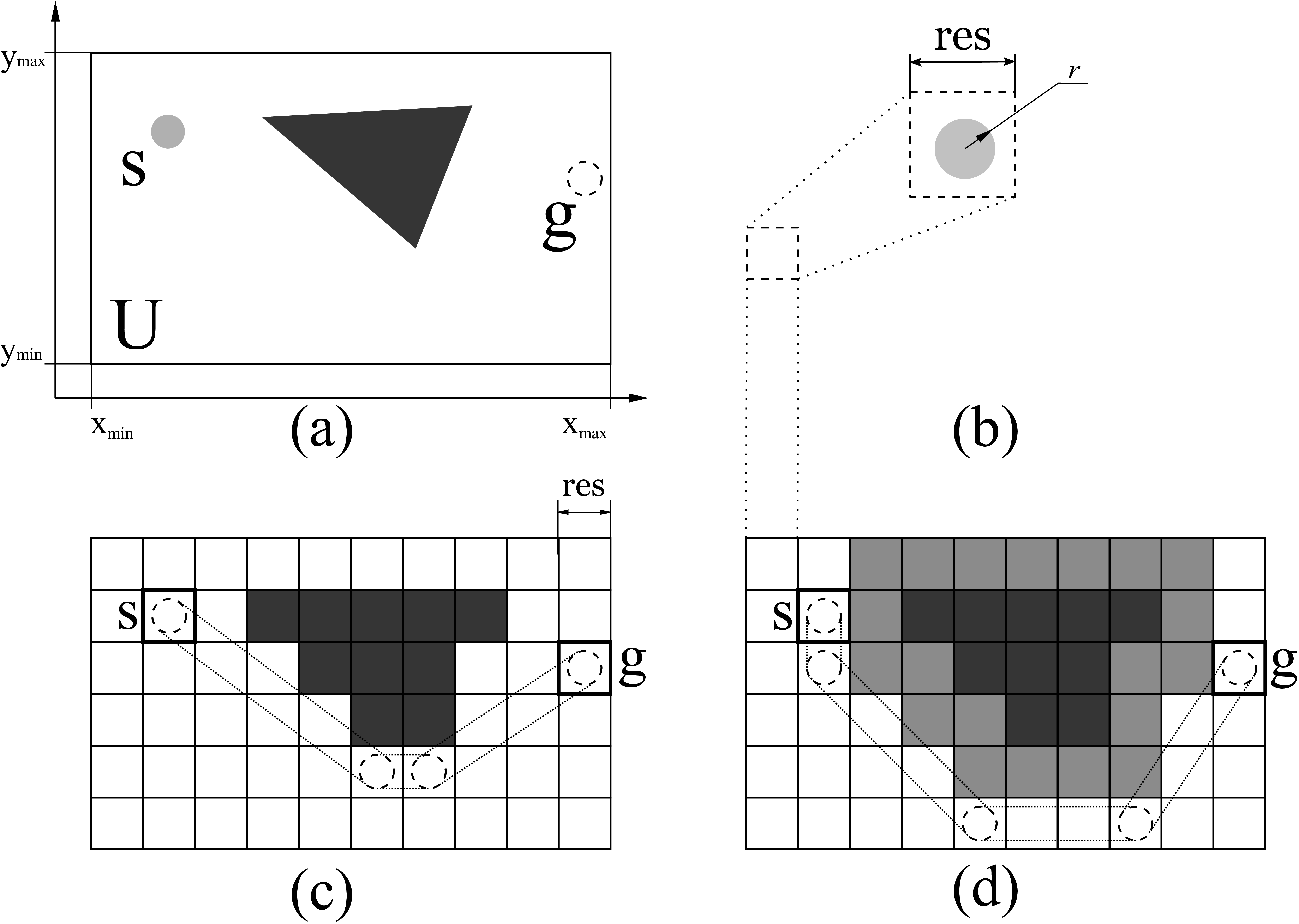}
	\caption{Grid representation of the workspace. a) Initial workspace. b) Square pattern used for discretization. c) Path on a grid colliding with the untraversable regions. d) Double-outlining of the obstacles and creating additional untraversable cells prevent from generating paths colliding with true obstacles.}
	\label{fig:grid}
\end{figure}

We suggest using grids as spatial representations for path planning as they are both informative and easy-to-search graph models of the agents' environment (as described in Section \ref{case}). Grid is constructed by overlaying regular square mesh over the workspace $U$ in such way that each grid element $c$, e.g. a cell, corresponds to a unique square area in $U$ sized $res \times res$, where $res$ --- is an input parameter. If this area overlaps with any obstacle, cell с is marked untraversable (traversable --- otherwise). We adopt center-based grid notation (in opposition to corner-based) meaning that a path should start (and end) at the center of some grid cell $c(x, y)$ (and thus it's supposed that any agent is tied to the center of some grid cell initially). We also adopt the idea of any-angle path finding \cite{Nash2007} and consider the path as the sequence of traversable but not obligatory adjacent cells $\pi=\{c_1, c_2, \dots, c_p\}$, such that an agent can move from one cell to next one in the sequence following the straight line connecting the centers of those cells. Function $los(c_i, c_j)\rightarrow\{true, false\}$ is given to check this condition. If the size of the cell is big enough to accommodate an agent (e.g. $res\geq 2r$) one can use well-known (and fast) Bresenham algorithm \cite{Bresenham1965} to check line-of-sight constraint. This algorithm identifies grid cells forming discrete representation of straight line, so after that, one needs to check if they all are traversable. Occasionally it can happen so that Bresenham algorithm identifies cells, that are all traversable, although actual straight line intersect an untraversable cell (and thus, possibly, an obstacle). To avoid this we suggest double-outlining the obstacles in the following way: after the grid is constructed mark all the adjacent cells for each untraversable and then put all the marked cells untraversable --- see Fig.~\ref{fig:grid}.

\begin{figure}
	\centering
	\includegraphics[height=3.5cm]{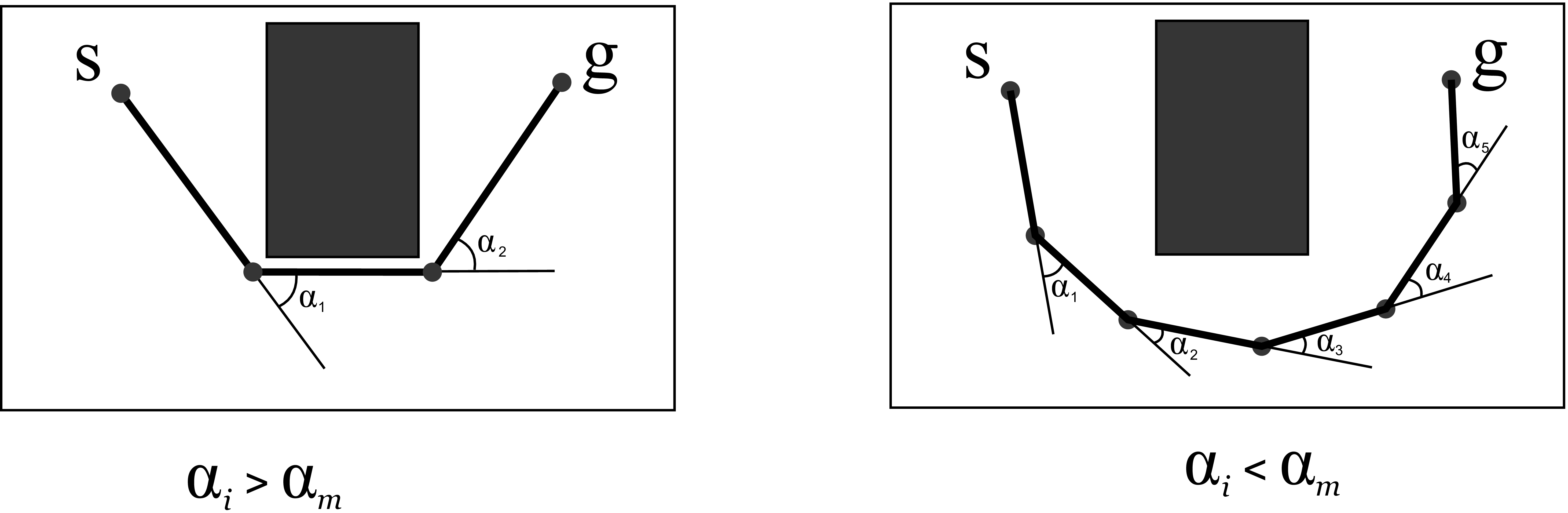}
	\caption{LIAN algorithm details.}
	\label{fig:lian}
\end{figure}

We use two algorithms for path finding: Basic Theta* \cite{Nash2007} and LIAN \cite{Yakovlev2015}. First algorithm searches for any angled path on given grid, second  searches for angle-constrained path, e.g. for such a path $\pi=\{c_1, c_2, \dots, c_p\}$ that an angle of alteration between any consecutive sections $\langle c_{i-1}, c_i\rangle, \langle c_i, c_{i+1}\rangle$ is less than the predefined threshold $\alpha_m$ (see Fig.~\ref{fig:lian}). Searching for the angle-constrained path is much more burdensome but such a path indirectly guarantees it's feasibility, e.~g. agent's ability to follow the trajectory in U defined by that path without violating the dynamic constraints (as described in Section~\ref{works}). 

So we search for any angle path first and if it is found start searching for the angle-constrained path. If it is found, we report success to the upper level, e.g. behavior planning module and wait for the next goal to be given. Failure to find a path means that angle constraint is too strict (in respect to the current environment model, e.g. grid, and start-goal locations) so the new sub-goal is to be given by the behavior planning module or the new angle constraint.

An interesting case occurs when any angle path planning returns failure, which means that there is no path to given goal location not due to the angle constraints, but due to the obstacles configuration, e.g. namely some obstacle is blocking the path (otherwise it would have been found as Theta* is sound and complete). In this case it is useless to ask for a new subgoal as a resultant path won't be found anyway. An obstacle blocking the path should be identified and its coordinates should be transmitted to the behavior planning module. To the best of authors' knowledge currently there are now works on the methods of identifying blocking obstacles, so this is an appealing research area to be investigated further.

Previously, when describing the path planning process we supposed that the goal cell is given, although, in the case sign-world model is used for behavior planning, the fuzzy goal area is under concern. This area is characterized by the point $cp(x, y)$ and radius $r_g$ --- if the agent reaches any point of the circle with radius $r_g$ and center in $cp(x, y)$ path planning is considered to be successfully accomplished. This can be taken into account in the following way --- execution of the LIAN algorithm should be stopped when any cell of the circle is under consideration. We also perform the consistency of the goal area check before path planning in the following manner: if given center point belongs to the grid's untraversable cell --- choose one of it's traversable adjacent cells as $cp$. If all the neighbors are untraversable --- examine their neighbors and so on up to the moment traversable cell of the goal area will be identified. If all cells forming the goal area are untraversable --- report  behavior planner and wait for the new goal area. 

\section{Behavior and Path Planning: case study}\label{example}

Below we demonstrate implementation of the planning method in solving the smart relocation task as described in Section~\ref{case}. We consider a simple case when two agents $A_1$ and $A_2$ form the coalition and share common goal area. Fragment of the agents knowledge base (sign model) is depicted in Tab.~\ref{tab:world}. Arrows used in description of image component split different columns of features (as described in Section~\ref{behavior}).

We consider the planning process of the agent $A_1$ whose goal situation is described with signs ``I - agent 1'', ``agent 2'', ``place $X_1$''. As said above all of these signs must be activated in knowledge base (sign word model) of agent $A_1$ which means that both ``I – agent 1'' and ``agent 2'' should be in ``place $X_1$''. Note that the description of $A_2$ goal situation is similar (``I - agent 2'', ``agent 1'', ``place $Y_1$''). Start situation for agent $A_1$ is depicted on Fig.~\ref{example} and is described by the signs ``I - agent 1'', ``place $X_4$'' $\rightarrow$  ``agent 2'', ``place $X_5$'' (here $\rightarrow$ is used to split different parts of situation description).

\setlength{\tabcolsep}{5pt}
\renewcommand{\arraystretch}{1.5}
\begin{table}\tiny
	\caption{The fragment of the sign world model of agents.}	
	\label{tab:world}
	\begin{tabular}{| p{1.5cm} | p{2.2cm} | p{2.2cm} | p{2.2cm} | p{2.2cm} |}
		\hline
		Specific signs of the $A_1$ agent
		&
		$s_1$: $n$=``obstacle 1''\newline
		$p$=\{``place $X_6$'', ``type $ot_1$''\}\newline
		$m$=\{``destroy 1''\}\newline
		$a$=$\emptyset$
		&
		$s_2$: $n$=``obstacle 2''\newline
		$p$=\{``place $X_2$'', ``type $ot_1$''\}\newline
		$m$=\{``destroy 1''\}\newline
		$a$=$\emptyset$
		&
		$s_3$: $n$=``far''\newline
		$p$=\{mechanical sensor parameters\}\newline
		$m$=\{``move 1''\}\newline
		$a$=\{``I move 1''\}
		&
		$s_4$: $n$=``agent 2''\newline
		$p$=\{mass, coordinates etc.\}\newline
		$m$=\{``send message''\}\newline
		$a$=\{``I send message''\}
		\\\cline{2-5}
		&
		$s_5$: n=``I move 1''\newline
		$p$=\{``I move 3'', ``I move 3'', ``place $X_1$''\}\newline
		$m$=$\emptyset$\newline
		$a$=$\emptyset$	
		&
		$s_6$: $n$=``I move 3''\newline
		$p$=\{``I'', ``here'' $\rightarrow$ ``empty'', ``place $X_3$''\}\newline
		$m$=\{``move 3''\}\newline
		$a$=\{path planning realization\}
		&
		$s_7$: $n$=``I --- agent 1''\newline
		$p$=\{mass, coordinates etc.\}\newline
		$m$=$\emptyset$\newline
		$a$=$\emptyset$	
		&
		$s_{10}$: $n$=``place $X_1$''\newline
		$p$=\{``far'', ``ahead'', ``right''\}\newline
		$m$=$\emptyset$\newline
		$a$=$\emptyset$
		\\ \hline
		&
		$s_{10}$: $n$=``move 1''\newline
		$p$=\{``I move 3'' $\rightarrow$ ``I move 3'' $\rightarrow$ ``place $X_1$''\}\newline
		$m$=$\emptyset$\newline
		$a$=$\emptyset$
		&
		$s_{10}$: $n$=``destroy 1''\newline
		$p$=\{``obstacle 1'' $\rightarrow$ ``empty''\}\newline
		$m$=$\emptyset$\newline
		$a$=$\emptyset$
		&
		$s_{10}$: $n$=``move 3''\newline
		$p$=\{``here'' $\rightarrow$ ``empty'', ``place $X_3$''\}\newline
		$m$=$\emptyset$\newline
		$a$=$\emptyset$
		&
		
		\\ \hline
		Specific signs of the $A_2$ agent
		&
		$s_1$: $n$=``obstacle 1''\newline
		$p$=\{``place $Y_2$'', ``type $ot_1$''\}\newline
		$m$=\{``destroy 1''\}\newline
		$a$=$\emptyset$
		&
		$s_2$: $n$=``obstacle 2''\newline
		$p$=\{``place $Y_3$'', ``type $ot_1$''\}\newline
		$m$=\{``destroy 1''\}\newline
		$a$=\{``I destroy 1''\}
		&
		$s_3$: $n$=``far''\newline
		$p$=\{mechanical sensor parameters\}\newline
		$m$=\{``move 1''\}\newline
		$a$=\{``I move 1''\}	
		&
		$s_8$: $n$=``agent 1''\newline
		$p$=\{mass, coordinates etc.\}\newline
		$m$=\{``send message''\}\newline
		$a$=\{``I send message''\}
		\\\cline{2-5}
		&
		$s_5$: $n$=``move 1''\newline
		$p$=\{``I'', ``here'' $\rightarrow$ ``empty'', ``place $Y_1$''\}\newline
		$m$=\{``move 1''\}\newline
		$a$=\{path planning realization\}	
		&
		$s_9$: $n$=``I --- agent 2''\newline
		$p$=\{mass, coordinates etc.\}\newline
		$m$=$\emptyset$\newline
		$a$=$\emptyset$
		&
		$s_{10}$: $n$=``place $Y_1$''\newline
		$p$=\{``far'', ``ahead''\}\newline
		$m$=$\emptyset$\newline
		$a$=$\emptyset$	
		&
		$s_{11}$: $n$=``destroy 1''\newline
		$p$=\{``obstacle'' $\rightarrow$ ``empty''\}\newline
		$m$=$\emptyset$\newline
		$a$=$\emptyset$	
		\\\hline
	\end{tabular}
\end{table}

PMA-procedure executes its first iteration with the start situation and goal sit-uation as input arguments. The M-step searches for the maximal effect-covered procedural feature from the set of available significances (``move 1'', ``send mes-sage''). From the Tab. 1 we can find that it is significance m=''move 1'' (transition ahead and right). Then the A-step is executed and in our case the procedure Ξ gives the personal meaning that includes feature corresponding ''I move 1''.

\begin{figure}
	\centering
	\includegraphics[height=6cm]{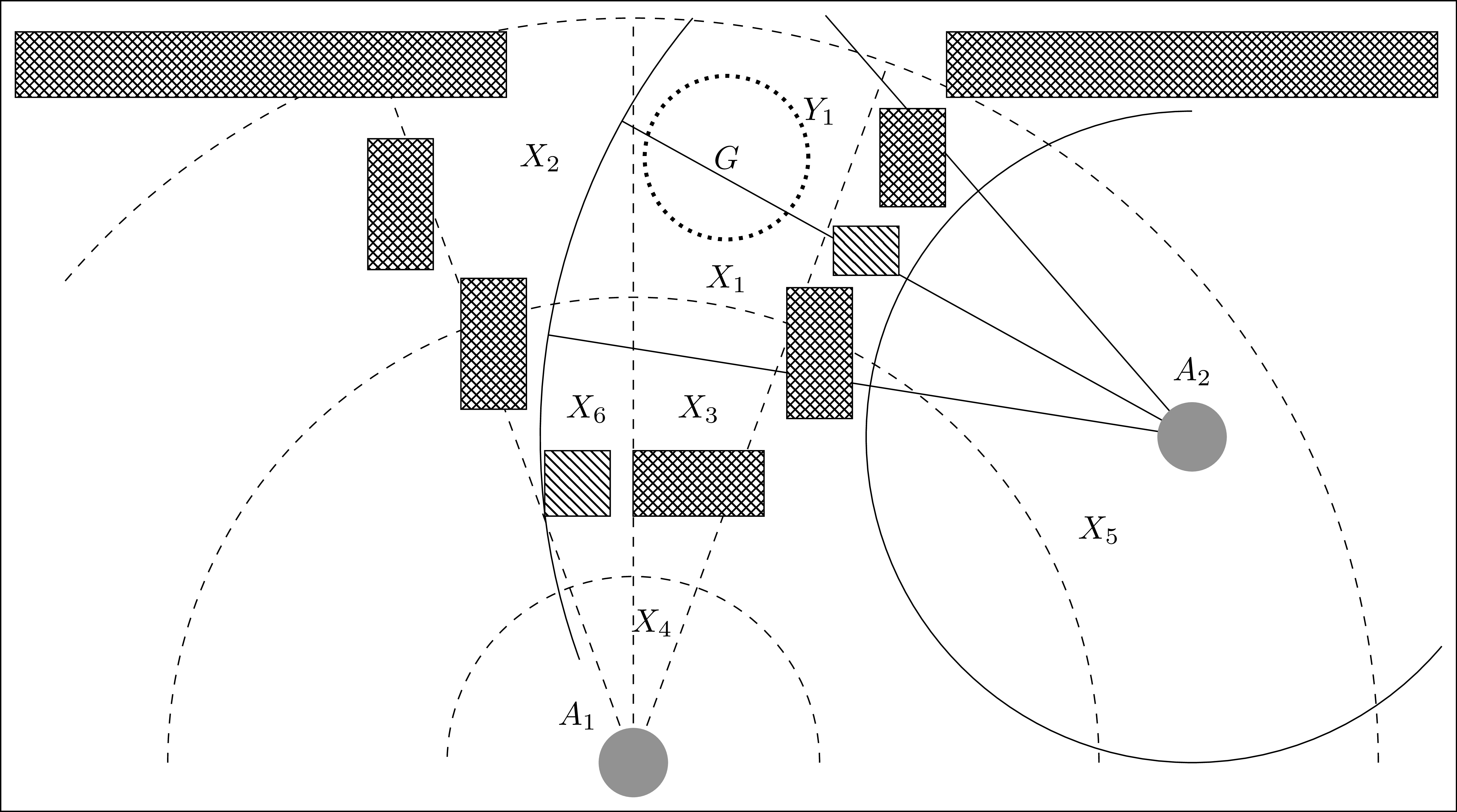}
	\caption{Representations of space in sign level. $X_i$ - names of places associated with signs for the agent $A_1$, $Y_1$ - a name of the place associated with the sign for the agent $A_2$.}
	\label{fig:example}
\end{figure}

Suppose agent $A_1$ has such a cognitive quality (introspection level) used as in-put parameter for behavior planner that leads to S-step of PMA procedure. On the S-step activation spreads top-down in procedural features hierarchy  and the activation trace (see Tab. 1) is ``I move 1'' $\rightarrow$ ``I move 3'' $\rightarrow$ ``path planning pro-cedures''). So finally path planning process is invoked. On given map path can not be found as ``obs 1'' is blocking the way, so it’s coordinates are returned back to behaviour planner. These coordinates are low-level image features so the im-age-base hierarchy is traced bottom-up until the sign mediating the obstacle will be reached (activated), e.g. ``obstacle 1'' sign. This sign is added now to the de-scription of start situation: ``I - agent 1'', ``place $X_4$'' $\rightarrow$  ``agent 2'', ``place $X_5$'' $\rightarrow$ ``obstacle 1'' and the description of final situation: ``I - agent 1'', ``agent 2'', ``place $X_1$'' $\rightarrow$ ``obstacle 1'', ``empty''. The PMA-procedure re-executes on the newly defined start situation. Again at the M-step the search for the maximal effect-covered procedural feature from the set of available significances (``move 1'', ``send message'', ''destroy 1'') is done and procedural feature ``destroying 1'' is selected. The A-step fails to fine suitable personal meaning (as agent $A_1$ has no destroying action). Thus communication action will be selected. So $A_1$ sends message to the agent $A_2$ with coordinates of obs1 and description of the action (``destroy 1''), needed to be performed in order for the agent $A_1$ to achieve the goal (``agent 1'',''place $X_1$''). The presence of the agent $A_1$ in the goal area (``agent 1'',''place $X_1$'') is required for $A_2$ to reach it’s goal (``I - agent 2'', ``agent 1'', ``place $Y_1$'') so signs ``obstacle 1'' and ``destroy 1'' (contained in transferred message) are added to $A_2$ goal which becomes: ``I - agent 2'', ``agent 1'', ``place $Y_1$'' $\rightarrow$ ``obstacle 1'', ``empty''. As $A_2$ has personal meaning ``I destroy 1'' A-step of the PMA-procedure for $A_2$ will be accomplished with success and the obstacle will be de-stroyed. So both of the agents can now construct the plan of goal area achieve-ment and thus accomplish smart relocation task.

\section{Conclusion}

In this paper we have presented an approach to build a two-layered planner for the robotic system which is a part of a coalition solving common task. We suggest following semiotic approach to develop cognitive top-level planner in order to make the system more versatile, robust and human-like (in contrary to the existing logic based approaches --- PDDL languages etc.). We have introduced new knowledge representation formalism --- sign world model --- which aligns well with the results of recent cognitive and neurophysiologic research. Path planning operators are the integral part of this hierarchical model so task and path planning processes are tied together as a parts of coherent framework. Behavior planning is a recursive search process in the hierarchical state-space induced by sign representation ending up with path planner triggering: when behavior planning reaches the lowest level of sign world model state-of-the-art grid-based planners are executed. We suggest focusing on searching for the special type of smooth paths --- angle-constrained paths --- in order to indirectly satisfy agent's dynamic constraints and produce feasibly trajectories. We also discuss involving sound and complete any angle path planner in the loop in order to be able to better distinguish between different failure outcomes of the path planning process.

We believe that proposed approach leads not only to extensive flexibility of the planning system which is now capable of solving collaborative navigation tasks which are not solvable by the individual members implementing traditional path planning algorithms, but also will significantly aid solving various human-robot interaction problems. One of worth mentioning tasks which can be positively impacted is natural language formulation of collaborative task to the group (coalition) of robots. Regarding path planning, the task of the blocking areas identification (in case of planner failure) is an appealing direction of further research.

\subsubsection*{Acknowledgments.} The reported study was supported by RFBR, research projects No. 14-07-31194 and No. 15-37-20893.

\end{document}